\documentclass[sigconf, screen]{acmart}

\copyrightyear{2021}
\acmYear{2021}
\setcopyright{acmlicensed}\acmConference[FDG'21]{The 16th International Conference on the Foundations of Digital Games (FDG) 2021}{August 3--6, 2021}{Montreal, QC, Canada}
\acmBooktitle{The 16th International Conference on the Foundations of Digital Games (FDG) 2021 (FDG'21), August 3--6, 2021, Montreal, QC, Canada}
\acmPrice{15.00}
\acmDOI{10.1145/3472538.3472589}
\acmISBN{978-1-4503-8422-3/21/08}

\begin{document}

\title[TaikoNation]{TaikoNation: Patterning-focused Chart Generation for Rhythm Action Games}


\author{Emily Halina}
\affiliation{%
  \institution{University of Alberta}
  \city{Edmonton}
  \country{Canada}}
\email{ehalina@ualberta.ca}

\author{Matthew Guzdial}
\affiliation{%
  \institution{University of Alberta}
  \city{Edmonton}
  \country{Canada}}
\email{guzdial@ualberta.ca}

\renewcommand{\shortauthors}{Halina and Guzdial}
\begin{abstract}

Generating rhythm game charts from songs via machine learning has been a problem of increasing interest in recent years. 
However, all existing systems struggle to replicate human-like patterning: the placement of game objects in relation to each other to form congruent patterns based on events in the song. 
Patterning is a key identifier of high quality rhythm game content, seen as a necessary component in human rankings. 
We establish a new approach for chart generation that produces charts with more congruent, human-like patterning than seen in prior work.
\end{abstract}





\maketitle

\section{Introduction}

%%Par 1. Is about situating the area that this research is focused on, and identifying a problem that exists in that area. You want to start broad then get more specific. 
%%You might want to start with ``PCGML'' and then move to talking specifically about chart generation.
Procedural Content Generation via Machine Learning (PCGML) is defined as the generation of novel game content through machine learning models that have been trained on previously existing game content \cite{summerville2018proceduralcg}. 
Previously, many PCGML approaches have been applied to 2D side-scrolling platformer games, where they were used to generate new levels based on existing training data. 
However, many underexplored game genres exist that may benefit from PCGML approaches.
In this paper we focus on one such game genre: rhythm games. 
Rhythm games challenge the player to hit a series of predetermined inputs in time to a given song within a level. 
These levels, known as charts, are traditionally handcrafted by individual authors,  requiring both expertise and several hours of work to complete.
%% rockband sentence
Games such as Rock Band 4 have over 1700 charts available as downloadable content \cite{knoop_2016}, meaning this quickly becomes a large-scale problem.
%%(Another sentence here on how this takes specialized design knowledge and so is inaccessible to many people)
Chart creation requires specialized design knowledge that can take several months to hone and develop \cite{johannsen_2016, 5argon_2018}. 
This makes chart creation inaccessible to many people who wish to see their favourite songs charted. 
We focus on the problem of procedural chart generation via machine learning, generating a novel chart based on a given input song with a machine learned model.
A trained model for chart generation could help to make chart creation more accessible to a wider audience, and speed up the chart creation workflow for developers.

%%Par 2. Explain to the reader why the problem you identified in par 1 is not yet solved. Lots of ways to make this argument, but the important thing is to convince the reader this is challenging and/or that it represents novelty. 
Chart generation has been attempted in prior work, but there are open problems that remain to be solved.
For example, there is the problem of onset detection, which refers to analyzing a piece of audio to locate the beginning of each musical note or beat within it. 
Onset detection has been the central focus of prior work involving chart generation, notably Dance Dance Convolution \cite{donahue2017dance}. However, just placing notes at appropriate times is not sufficient to create a compelling, engaging chart, especially at a high difficulty level \cite{johannsen_2016}. 
For example, high level charts include ``patterning,'' the placement of game objects in relation to each other to form congruent patterns based on events in the song \cite{5argon_2018}.
Patterning is a fundamental component of highly-rated charts that allows individual authors to express their unique interpretations of a song \cite{5argon_2018}. 
%Patterning adds a new dimension to chart creation, allowing individual authors to express their unique interpretations of a song \cite{5argon_2018}. 
The problem of training a model to recognize the appropriate times to place notes to create consistent, human-like patterns is still largely unexplored.

%%Par 3. Okay, here's what we actually did, and make clear how this work addresses or gets us closer to solving that problem (even though it is very unlikely to have solved it solved it completely).
In this paper, we focus on the task of chart generation for the game Taiko no Tatsujin (abbreviated as Taiko) \cite{tpgpl_semro_megaapplepi_2020}.
Taiko is a long running rhythm game franchise modelled on simulating the playing of a taiko drum.
We chose Taiko due to its heavy emphasis on percussive rhythm, which made it a natural fit to approaching the problem of generating charts containing human-like patterning. 
After curating a dataset of 100 highly rated charts, we trained a Long Short Term Memory Recurrent Neural Network (LSTM RNN) to translate music to Taiko charts.
Unlike prior work \cite{donahue2017dance,lin2019generationmania, liang2019procedural}, we predict multiple outputs simultaneously, biasing the model towards longer-form patterns.
We call our system ``TaikoNation,'' a portmanteau of the name of a popular osu!Taiko mascot and ``generation.''
%We call our system ``TaikoNation'' as an allusion to a popular osu!Taiko mascot combined in a portmanteau with ``generation.''
Our system creates more congruent, human-like patterning than seen in prior chart generation work.

%%Par 4. Walk the reader through the rest of the paper in terms of its structure, and clearly state the contributions (the specific new and valuable ideas we're putting forward) of this paper. 

In this paper, we present the following contributions:

\begin{itemize}
    \item An LSTM architecture for generating novel Taiko charts based on arbitrary audio.
    \item A curated dataset of 110 Taiko charts along with their corresponding song data in a novel format suitable for machine learning. We also include 10 Dance Dance Revolution charts in this format used for comparison.\footnote{\url{https://github.com/emily-halina/TaikoNationV1}}
    \item Experiments comparing our model to an existing approach: Dance Dance Convolution \cite{donahue2017dance}.
\end{itemize}

\noindent
We begin by examining prior works involving chart generation and other relevant topics. 
Following this, we cover a full system overview detailing the architecture used for the model and the steps taken to process the input and output. 
We define evaluation methods for chart generation models centered around both onset detection and different notions of patterning, and compare our model's results against Dance Dance Convolution \cite{donahue2017dance}.
Finally, we discuss the future potential uses of the model beyond this paper in a co-creative context.

\section{Related Works}

%Pure novelty argument section, this is a very defensive section.

%Each subsection/paragraph should introduce an area of related work (so for example Rhythm Action Game Chart Generation, but you might also have sections on music generation, or ML tools, etc.). Then you'd talk about the general strategies or types of work in this area (without getting into detail, but still citing them). Then identify the 1-3 most similar examples of prior work, and describe them in 1-2 sentences each (only describe the aspects related to your work). Then end off by stating how your work differs still from even this most similar prior work.
\subsection{PCGML Approaches for Sequence Generation in Games}
%(Introduce PCGML again)
%(And then cover some examples of using sequence to sequence models (like LSTMs) for PCGML.
%Including Summerville's prior work on Mario and Magic the Gathering. 

PCGML, the generation of new content for a game through machine learning models trained on existing game content, has been applied to sequence generation problems like ours in the past. 
Many prior works have made use of sequence-to-sequence models such as LSTMs, which are commonly used for sequence generation problems \cite{summerville2018proceduralcg, hochreiter1997long, sutskever2014sequence}. 
LSTMs have been used to generate many different types of content outside of rhythm game charts like Super Mario Bros. levels and Magic the Gathering cards \cite{SMBRNN, summerville2016mystical}. 
We similarly make use of an LSTM trained for our task of Taiko chart generation.

%talking about generating smaller subsections and putting them together
A common approach across PCGML is to generate subsections of the final desired content in order to better leverage existing data \cite{snodgrass2016controllable, sarkar2020sequential}.
This approach has been used in multiple game domains to generate subsections of levels which are then strung together  \cite{thakkar2019autoencoder, schrum2020interactive}.
We similarly make use of this approach, but focus on generating longer subsections than prior work in PCGML chart generation \cite{donahue2017dance, liang2019procedural, lin2019generationmania}. 

\subsection{Chart Generation \& Music-Related PCG without Machine Learning}
%Talk about chart generation/music-related PCG without ML Control F rhythm in this paper https://scholar.google.com/scholar?hl=en&as_sdt=0%2C5&q=rhythm+games+generation+PCG&btnG=
Early attempts at chart generation for rhythm games utilized rule-based techniques \cite{o2003dancing} and genetic algorithms \cite{nogaj2005genetic} to determine where to place game objects within charts. There is also prior work utilizing PCG techniques with a rhythmic focus for other game domains, such as platformers \cite{smith2009rhythm, smith2012pcg}. 
Prior work also includes PCG based on user input, altering or blending songs together based on player actions \cite{gillian2009scratch, jordan2012beatthebeat}. There are also several commercial games focused around the use of PCG techniques on arbitrary music to generate content \cite{beathazard, audiosurf}. 

\subsection{Chart Generation with Machine Learning}
There have been a number of instances of prior work on the topic of PCGML chart generation for rhythm games. 
These approaches break the chart generation pipeline into multiple distinct segments and handle the problem in pieces. 
Donahue's Dance Dance Convolution \cite{donahue2017dance} utilized two separate machine learning models to handle the tasks of note placement (deciding when to place a note in the song) and step selection (deciding which input to assign to each note) respectively. 
By note we indicate a given musical onset that is assigned a game object that requires player input.
Liang's work Procedural Content Generation of Rhythm Games (PCGoRG) \cite{liang2019procedural}, further extended the model used in Dance Dance Convolution, improving the onset detection model through the usage of larger stride windows (called ``fuzzy labelling''). 
Lin's GenerationMania \cite{lin2019generationmania} focused on sample classification and selection. 
This additional focus is due to the nature of the game domain for Generationmania: Beatmania IIDX, which is a ``keysounded'' game. In keysounded games, the player's input directly influences the song, as each note directly represents a musical sample within the track. 
This causes the player to generate the song live as they play the chart, which adds a new dimension to the problem of chart generation for IIDX. 
In contrast, Taiko is a non-keysounded game, so this problem is not present within our work.

There are multiple standout differences between our approach and these three prior attempts at chart generation. 
Notably, in these examples of prior work, the problem of chart generation was divided into multiple parts, with onset detection and game object selection being split into separate problems. 
In our model, the two problems are handled simultaneously with the goal of creating a stronger link between the timing and object type of each given note.
In addition to this, each of the three prior approaches have a built in notion of difficulty, attempting to adjust the density of a given output chart based on the desired difficulty. 
In contrast, our work focuses on the expert level of difficulty, because of our focus on the more complex relationships and patterns present between notes at a high difficulty. 
In general, PCGML approaches have proven less effective for lower difficulties so far, as valid low level charts have substantially greater sparsity. 
While these examples of prior work have a notion of prior context due to implementations of systems like ``chart summaries,'' which contain selected information from the previous measures of a chart, they still predict output for each timestep individually. 
Our model has a sense of structure and summary embedded into it, as we predict over multiple timesteps worth of input at a time.
When combined, we hypothesize that these differences will lead to a better handling of patterning than these prior approaches, leading to more congruent placement of notes in relation to each other.

%%(One last sentence here that brings up patterning again, and indicates that all of this together should lead to a better handling of patterning than these prior approaches).

\section{System Overview}

%There has to be sufficient detail in this section that someone who doesn't read anything but this paper should be able to reimplement your work.

%Figure to explain the process and a figure to explain your model

%%\begin{figure}
   %% \centering
  %%  \includegraphics[width=\columnwidth]{EmilyPCGWorkshop/method-overview-32.png}
  %%  \caption{Caption}
  %%  \label{fig:my_label}
%%\end{figure}

%%As can be seen in Figure \ref{fig:my_label}.
% change visualization for input / output

% hello, currently fixing the figures so they aren't squished by the 2 columns
\begin{figure*}
    \centering
    \includegraphics[width=\linewidth]{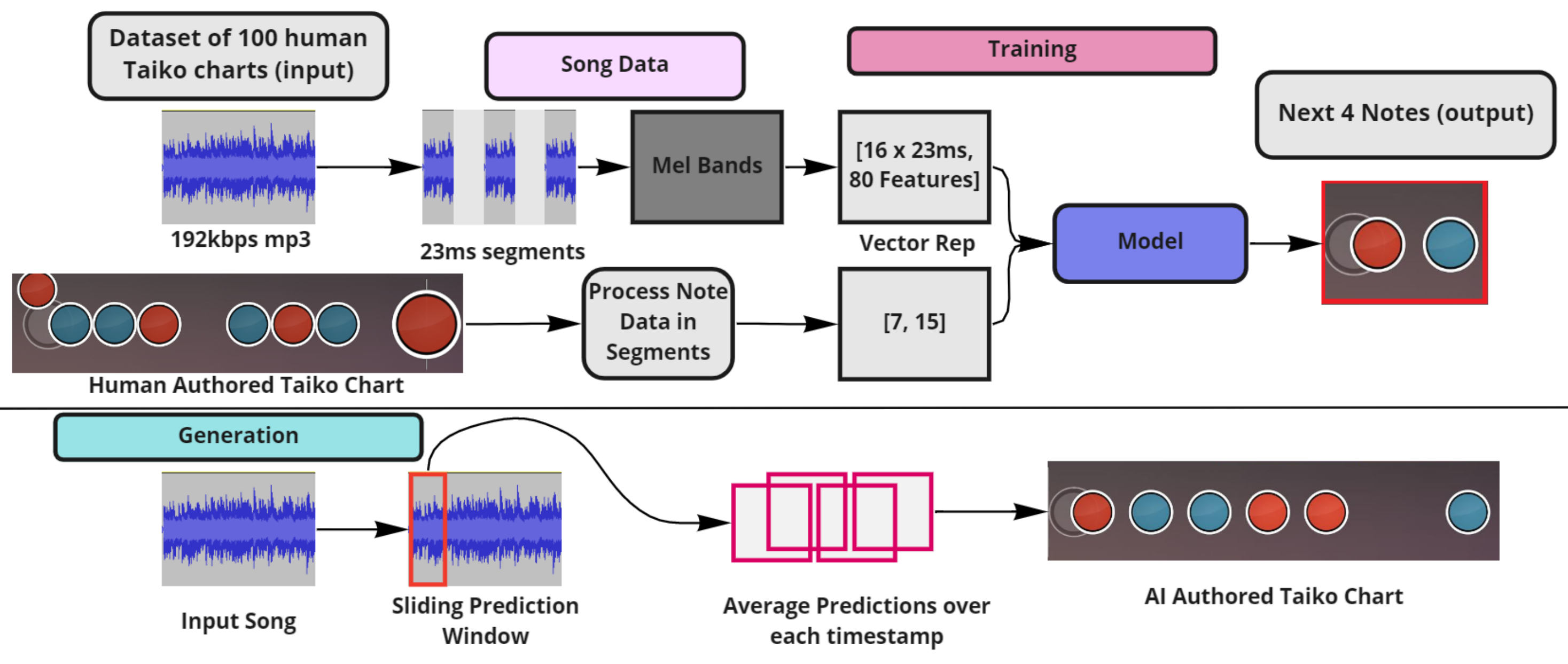}
    \caption{System Overview of our chart generation pipeline. We use a LSTM DNN architecture trained on a curated, human-authored dataset to predict the next four notes of a chart from a combination of song and note data. We use this model to make predictions on a given input song over a sliding window, averaging the predictions for each timestamp to output a chart.}
    \label{fig:sysover}
\end{figure*}

% start by walking through the main figure
Figure \ref{fig:sysover} depicts the chart generation pipeline from the initial dataset to the creation of a Taiko chart based on arbitrary song input. 
%Some sentence here indicating that you'll walk through the figure now and go into more detail on each part later in the section.
After giving a broad overview of the pipeline, we will go into detail in the following subsections.
We begin with a curated dataset of 100 Taiko charts, with song data in the form of a 192kbps mp3 audio file and note data contained in a .osu file \cite{osufile} for each chart. 
We slice the audio file into 23 millisecond (ms) segments and extract the audio features from each segment.
We parse and convert the note data from the .osu file into a new representation for training, which we describe below.
This pre-processed data is fed into our LSTM DNN architecture in bundles of 16 segments at a time, and the model predicts the next 4 notes of the chart based on the previous note and chart data.
After training on the dataset, we use this model to generate predictions on a sliding window of 16 x 23 ms across a given input song, averaging the predictions for each timestamp together. 
These averaged predictions are converted back into a playable format with some light post-processing, creating a chart.

\subsection{Human-Like Patterning}
% "formally" introduce patterning
Our system attempts to replicate the notion of ``human-like patterning,'' or shortly ``patterning.'' 
Patterning is not a well-defined term, and we are not attempting to define it here. 
Instead, we give our interpretation of patterning and provide some examples of what patterning entails in the context of this paper.

% define what we will be considering then give examples
The main aspect of patterning considered in this paper is the placement of game objects (notes) in relation to each other to form congruent patterns based on events in the song \cite{5argon_2018}.
As an example, three notes placed directly next to each other on the timeline form a triplet, which is a specific rhythmic pattern.
Patterns can occur on multiple scales, from over a particular measure to over an entire chart.
For example, a specific recurring set of notes over the course of a song could be considered a higher-level pattern, analogous to a motif in music.
The congruence between notes within patterns can be based on rhythmic placement as well as the types of notes used.
In Taiko, these two congruences are highly interconnected, with the majority of patterns found in high quality human-authored charts utilizing both note type and rhythmic structure simultaneously.
For instance, chart authors may use the same rhythmic structures multiple times in a row with different note types.
This can lead to patterns that feel completely different to play despite having the same rhythmic structure.
We focus mainly on replicating the rhythmic half of this problem while also attempting to match the appropriate distribution of each type of note object found in human-authored charts.
%With these two aspects of patterning at the forefront of our system's design, we aim to better replicate and recreate the patterns found in high quality human-authored charts.

\subsection{Data}
% paragraph about how the dataset was collected
%(Option 1: Sentence here saying that our goal is to investigate patterning and thus we need high quality charts that showcase intricate patterning.)
Our goal is to investigate patterning, and thus we need a collection of high quality charts that showcase intricate, song-appropriate patterning.
The dataset was collected from a community run database of approved charts for Taiko \cite{herbert_2007}.
These charts all follow a set Ranking Criteria \cite{rankcrit}, requiring them to be scrutinized by experienced creators before they are approved.
Since these reviewers value patterning \cite{rankcrit}, these charts are more likely to contain the high quality patterning we are attempting to model.
We further verified this was the case based on an author's expert knowledge of patterning.
We sorted these approved charts by user rating from high to low, and took any chart above a certain difficulty threshold.
This threshold was set at a relatively high level of difficulty to ensure the dataset contained a varied collection of dense, interesting patterns.
If there was more than one chart for the same song that was above this difficulty threshold, the higher rated of the two was chosen. 
For this initial exploration, we collected 100 charts with a 90-10 split between the training and validation set.
This resulted in approximately 1.3 million elements to train on once processed.
These 100 charts represent a large number of the database's top rated charts, and cover a wide span of genres and charting styles.

% paragraph about how the song data was processed
Our approach to audio representation is similar to Dance Dance Convolution's \cite{donahue2017dance, schluter2014improved}, with a few differences based on game domain specifications and our focus on patterning.
Each chart's corresponding audio was initially a 192kbps mp3 file.
The specific audio quality is due to Ranking Criteria specification \cite{rankcrit}.
While there is a concern for audio artifacts due to this low bitrate, each audio file in the dataset has been reviewed by multiple chart authors.
This ensures the best possible audio quality as a part of the aforementioned ranking process.
We cut the audio into 23ms segments, converting each segment to a monaural, or single channel, .wav file for processing.
23ms segment sizes were chosen after an analysis of note placements relative to beats per minute (BPM) within the dataset. 
Specifically, 23ms is roughly the distance between two 1/64th notes at 163 BPM, and offers a reasonable divide for most BPMs outside of extreme ranges.

We perform a short-time Fourier transform (STFT) on each of our extracted segments, using a window and stride length of 23ms to match our segment length.
% justification
STFTs are a fundamental tool of signal analysis, and have been used in onset detection frameworks similar to ours in prior work \cite{allen1977unified, schluter2014improved}.
While Dance Dance Convolution used a multiple-timescale transform for its onset detection pipeline, we instead used only one timescale length. 
This is because we use a larger stride length, which along with appending the previous note data provides the rhythmic context that would otherwise be given by using multiple longer timescales \cite{hamel2012building}.
We then compress the dimensionality of our spectra down to 80 frequency bands by applying a Mel-scale filterbank \cite{stevens1937scale}.
This allows us to extract the key features from the STFT results into evenly spaced divisions, which are then scaled logarithmically to account for human perception of loudness \cite{donahue2017dance}.
We used the ESSENTIA audio processing library \cite{bogdanov2013essentia} for this task, normalizing each frequency band to zero mean and unit variance.
After this processing, we append the previous 15 segments of audio data to each segment, giving a dimensionality of (16 x 80) per segment, representing 368ms of song data. 

% paragraph about how the note data was processed
The chart information from the dataset comes in the form of .osu files, a human-readable file format containing information about the game objects used in the chart and their timestamps \cite{osufile}. 
\begin{figure}
    \centering
    \includegraphics[width=\columnwidth]{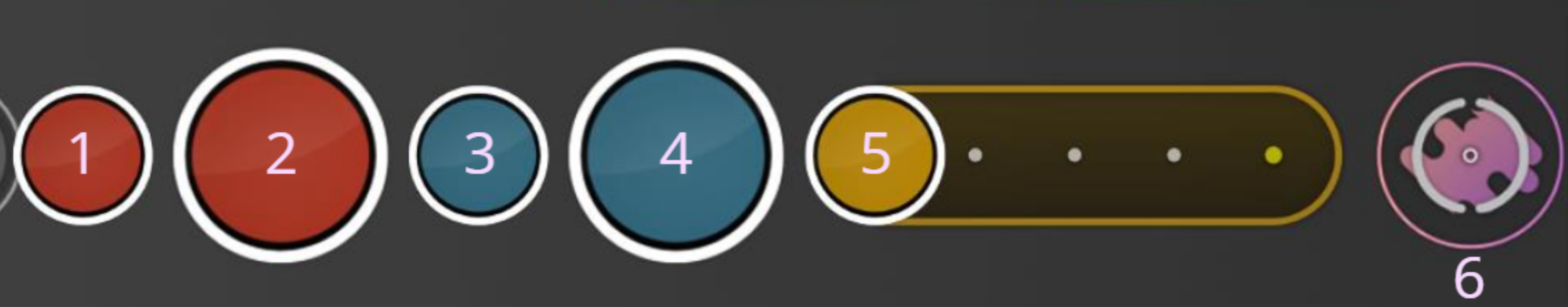}
    \caption{Depiction of each note type present within Taiko. In order these are: small Don, big Don, small Kat, big Kat, Drumroll, and Denden.}
    \label{fig:object}
\end{figure}

We parsed this file format, extracting the note data in 23ms timesteps and encoding the various game objects into a 7-length one-hot representation.
This representation included an entry for ``no note,'' along with all 6 of the main object types found in Taiko. These are:
\begin{itemize}
    \item Small / Big Don: 1 \& 2 in Figure \ref{fig:object}
    
    One of two main types of notes. Dons have two distinct keys (or the center of the drum) associated with them. Small Dons require you to hit one key, and big Dons require you to hit both keys.
    
    \item Small / Big Kat: 3 \& 4 in Figure \ref{fig:object}
    
    The second of two main types notes. Kats have two distinct keys (or the rim of the drum) associated with them. Small Kats require you to hit one key, and big Kats require you to hit both keys.
    
    \item Drumrolls: 5 in Figure \ref{fig:object}
    
    A rare object type that requires drumming in time to a consistent 1/4 rhythm for a given duration. Any keys or area of the drum can be used during these segments.
    
    \item Dendens: 6 in Figure \ref{fig:object}
    
    A rare object type that requires the player to hit the keys / drum a set number of times before time is up. These are not associated with the rhythm of the song, and can be considered analogous to ``drum fills''.
\end{itemize}

\noindent
At the timestamps where the model should predict a note, we represent that note by filling the index of the given note vector with 1s.
This led to a dimensionality of (7 note features x 15 23ms timestamps) for each segment of note data. 

\subsection{Architecture}

\begin{figure*}
    \centering
    \includegraphics[width=\linewidth]{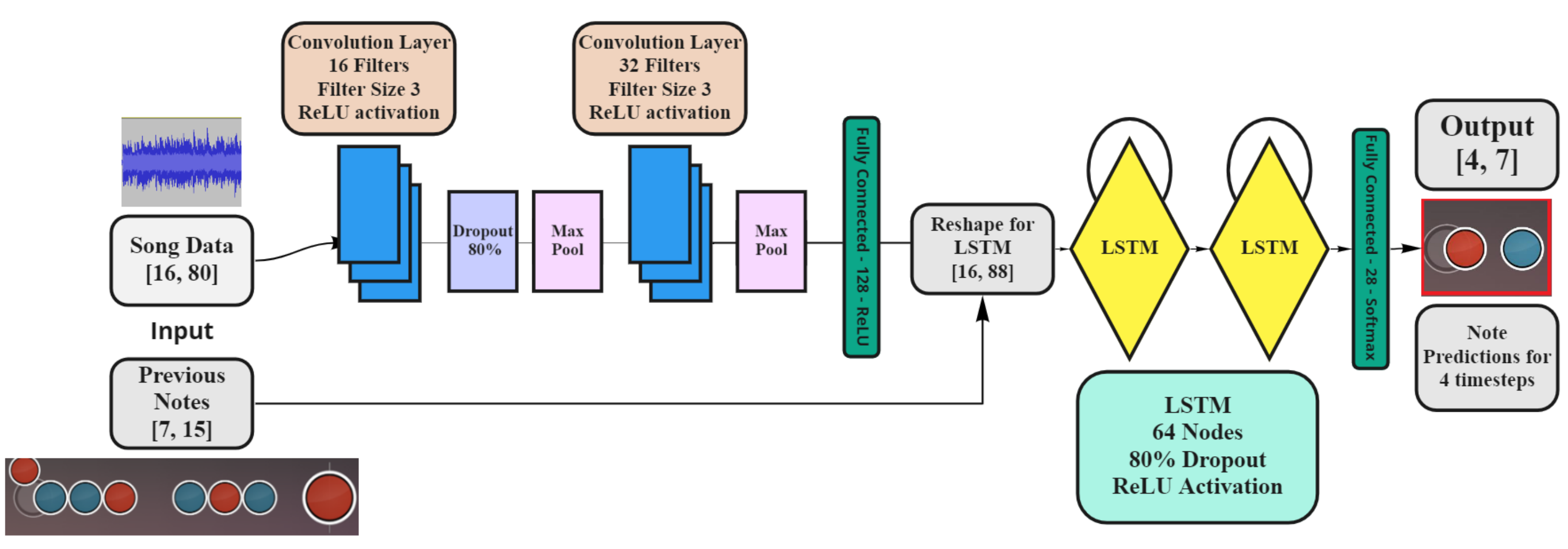}
    \caption{A visualization of our LSTM DNN Architecture. Two convolutional layers are used to extract features from the input segment of song data, which represents 368ms of the input song. These features are then combined with the previous note data and passed through two LSTM layers before outputting note predictions for the following four timestamps in the song.}
    \label{fig:arch}
\end{figure*}

% walking through the figure
A visualization of our LSTM DNN architecture can be seen in Figure \ref{fig:arch}.
The input consists of 16 segments of song data representing 368 ms of audio and 15 segments of the corresponding previous note data.
We run the song data through a convolutional layer with 16 filters using a rectified linear unit (ReLU) activation function.
Convolutional layers are commonly used to parse sound data for many signal processing tasks including onset detection \cite{lee2009unsupervised, schluter2014improved}.

%(We chose ReLU as it has been demonstrated to better model complex, nonlinear relationships (CITE))
ReLU has been demonstrated to better model complex, non-linear relationships than other activation functions \cite{xu2015empirical}.
For this reason we use ReLU activation for the remaining layers, excluding the final output layer.
This is followed by a 80\% dropout layer, then a max-pooling layer.
Dropout layers are helpful to reduce overfitting, allowing us to learn more general models \cite{srivastava2014dropout}.
The 80\% value arose due to a class imbalance, as roughly 83\% of our notes are the ``no note'' class.
Max-pooling layers were used as we wanted to encourage the network to focus on large onsets for note placement.
%(Dropout layers are helpful to reduce overfitting and learn more general models. The 80\% arose due to a class imbalance, the majority of our notes (83\%) were the no note class.)
%(Max-pooling was used as we wanted to encourage the network to focus on large onsets for note placement.)
This is followed by another convolution layer with 32 filters, followed by another max-pooling layer.

We pass this result through a fully-connected layer with 128 nodes, which is reshaped to an (8 x 16) vector.
This is done to combine the CNN output with the note data by performing by-element multiplication on each segment.
We chose a by-element multiplication instead of treating these as two separate channels in order to tightly associate the tasks of onset detection and object selection, which were treated separately in prior chart generation work \cite{donahue2017dance, lin2019generationmania, liang2019procedural}.
Since the final segment does not have note data as part of our input, we multiply that segment by a vector of 1s as a placeholder. 
%(Something on what you just said here. Basically, we wanted to tightly associate these two things together instead of treating them as separate channels). 

From here, we run our combined data through two LSTM layers with 64 nodes, 80\% dropout, and ReLU activation. 
LSTM layers were chosen due to their suitability for sequence-to-sequence learning tasks, and have been demonstrated to perform well on timing dependant audio tasks \cite{sutskever2014sequence, gers2002learning}.
Finally, we pass this data through a softmax activated fully connected output layer with 28 nodes.
%(Softmax activation is typically used when predicting one particular output class)
Softmax activation is typically used when predicting one particular output class, as it acts as a probability distribution across each of the possible output classes \cite{memisevic2010gated}.
This is reshaped to a (4 x 7) vector, representing predictions for which gameplay objects to place at the next 4 given timestamps.

% training details
We trained the model using the Adam optimizer and categorical cross-entropy as a loss function.
Taiko charts have a large amount of variance, as there can be several valid charts for the same song, which is why we chose Adam over SGD \cite{kingma2014adam}.
Initially, we trained the model for 10 epochs with a learning rate of 0.00001 and a batch size of 16.
These values were experimentally verified on our validation set. 
After this initial training, we dropped the batch size down to 1, turning down the learning rate to 0.000005 and retraining one epoch at a time until weight explosion, taking the prior epoch's model.
This process took 4 additional epochs.
While this is not the most robust training methodology, we employ it for this initial exploration of the approach due to the high variance in the data. 

In total, the model trained for approximately 2 and a half hours on a single machine using an AMD Ryzen 5 3600x CPU and NVIDIA GeForce 1080 Ti GPU.

\subsection{Generation}

% walk through the generation pipeline
Our chart generation pipeline begins with a full song as input. 
We begin by processing the song, slicing it into 23ms segments and converting it to our uniform monaural .wav file format.
Next, each slice of the song has its features extracted as per the processing described in Section 3.2.
After the song data has been prepared and processed, we use the model to make predictions on the song data using a sliding window of 16 segments, or 368ms.
We initially use placeholder note data for the first 16 segments, which takes the form of a vector of 0s. 
This is because there are very rarely notes in the immediate starting moments of the audio file, so we used the placeholder notes as a buffer for the model. 
We then feed each following prediction into a queue to be used in the following inputs.
There are 4 different predictions for each timestamp due to the structure of the output, which are each 7-length vectors.
We sum each of these prediction vectors together, normalizing the result.
This result is treated as a probability distribution, which we sample from for a final prediction for each timestamp.
% compare to just taking prob from most recent vs sum, more likely to select "most likely object" at that point rather than respecting overall long term patterning
We treat each result this way rather than taking the sample from only the most recent prediction in order to bias the system towards longer term structure.
This promotes the model to not just select the ``most likely object,'' and avoid discounting underrepresented classes which otherwise arise from the inherent class imbalance in the dataset.
%We treat each result this way to give stronger weightings more of a chance of being selected while not discounting the underrepresented classes which arose from the class imbalance.

%post processing and converting to playable format
These predictions are then fed into a post-processing script which converts them into a playable format. 
This post-processing step eliminates ``double positives'' by locating notes that are only 23ms apart. 
Notes are almost never this close together in human charts, so we remove the later of the two notes when this occurs, which helps to smooth out the resulting output chart. 

\section{Evaluation}

%Outline and argue as to why the evaluation that you used is appropriate

%hi!
%Remind the reader what we were trying to accomplish
Our goal is to create a chart generation model that leads to output with better patterning, defined as congruent combinations of notes. 
Deliberate patterning according to a given song is a task which requires expertise for human chart creators \cite{5argon_2018}. 
The task of patterning was divided into the separate tasks of onset detection and game object selection in prior work \cite{donahue2017dance, lin2019generationmania, liang2019procedural}. 
In our approach, we handled these two tasks within the same model with the goal of improving on patterning in comparison to prior works. 
After a brief overview of our evaluation methodology, we will go into detail regarding our baselines and the metrics we used.

%Give the basic task of the evaluation, which will be to produce charts for unseen songs for which we have human charts, and then we'll compare these according to a number of metrics
The basic task of the evaluation is the generation of charts for withheld songs. 
We generate charts for these withheld songs using both our model and two other baselines, Dance Dance Convolution \cite{donahue2017dance} and random noise. 
These baselines are elaborated on below.
We convert all the charts into a binary format for comparison.
This conversion involved stepping through each song 23ms at a time, representing any timestamp with an object that requires a discrete input with a 1, and everything else with a 0. 
We defined these discrete inputs to be any drum hit or beginning of a held object in Taiko, and any step or beginning of a held step in DDR.
The purpose of this conversion was to make the comparison between the two models as fair as possible, as well as to allow cross-game comparisons.

%Give the baselines that we used and why they're appropriate (also indicate why we didn't use baselines that might seem "obvious"). 
The baseline we chose from prior chart generation work involving ML is Donahue's Dance Dance Convolution (abbreviated DDC) \cite{donahue2017dance}.
We selected DDC for three major reasons: DDC's ubiquity as a baseline, the similarities in game domain between Taiko and Dance Dance Revolution (abbreviated DDR), and access to similar quality human-authored charts for comparison. 

DDC is a ubiquitous baseline for comparison within other prior PCGML chart generation works \cite{lin2019generationmania, liang2019procedural}. 
DDC's general focus on the task of chart generation provides the most clear baseline for comparison versus the two other mentioned prior works, which additionally focus on the separate tasks of sample classification \cite{lin2019generationmania} and improved onset detection methods \cite{liang2019procedural}. 
Our specific focus is on evaluating patterning, which is unrelated to these separate tasks found in other work.
Similarly, DDR and Taiko are non-keysounded games, which make them better candidates for comparison than Beatmania IIDX, the keysounded game domain of GenerationMania \cite{lin2019generationmania}.
This is due to the significant differences between charting for keysounded and non-keysounded games which arise from the inherent restrictions imposed by keysounding \cite{5argon_2018}.
Finally, we have access to a user-curated database of charts for DDR similar in form and function to our source of Taiko charts \cite{itgpacks}.
This allows us to find human charts for the same songs in both Taiko and DDR with appropriate difficulty levels for comparison that were reviewed and approved to be apart of a community database.

%Both DDR and Taiko are non-keysounded games, meaning objects do not directly correspond to individual notes in a song, where as IIDX is keysounded.

%While osu!mania is more similar due to the game's lack of mandatory keysounding, osu!mania has inputs associated with the player releasing a held key at a given time. 
%There is no game object within Taiko or DDR that is analogous to this ``release'' object, making direct comparison between our model and PCGoRG \cite{liang2019procedural} less viable than comparison with DDC.

After collecting an evaluation set of 10 human-authored charts for the same songs in both Taiko and DDR, we generated charts for each song using TaikoNation and DDC respectively.
It was necessary to collect separate human datasets for both Taiko and DDR due to differences in charting paradigms within each game domain.
We will be comparing each chart generation method to each of the human-authored datasets individually.
%-Extra Sentence here on why its necessary to collect the two different game domains, and make clear you'll be comparing each ML generator to each of the two human chart sets.-
%-We convert all charts to the same binary representationd described above, since the game objects/input controls differ between games-
We convert all of the charts to the same binary representation described above to account for mild discrepancies in game objects between DDR and Taiko.
We decided to select 10 charts for the evaluation set so that we could ensure that we covered an appropriate breadth of musical genres while ensuring both charts were of the appropriate difficulties and lengths to be fairly compared.
In the selection of the Taiko charts, we used the same difficulty threshold we identified in Section 3.2 for selecting charts for the dataset, employing a similar threshold for DDR charts.

Along with DDC, we also selected random noise as a comparison baseline to be used as a control.
We constructed this random noise by randomly choosing either 0 or 1 at each 23ms timestamp in the binary representation.
This is used to evaluate how much structure both models are imposing on their output in comparison to human created charts.
Noise also serves as a control for our chosen evaluation metrics, helping to justify that they are valid methods of measurement.

%Then give those metrics and explain why we're including these metrics
% vs random noise, vs human file, overall pattern space, overall intersection between human pattern space and ai pattern space

We make use of 5 metrics in our evaluation which involve comparisons against human charts to measure onset detection and patterning.
These metrics are:
\begin{itemize}
    \item Direct Comparison against Random (\textbf{DCRand}) 
    
    Compares the binary representation of the given charts against random noise, counting the one-to-one similarities at each timestamp. 
    This metric measures the AI output's distance from randomness to determine how much structure each approach is introducing to its output.
    \item Direct Comparison against Human (\textbf{DCHuman}) 
    
    A one-to-one measure of similarities per timestamp between the binary representations of the AI generated and human authored charts for a given song. 
    We chose this metric to give an overall picture of similarity between the given charts without a strict focus on onset detection.
    It is equivalent to accuracy, if we consider the human chart to be the gold standard. 
    \item Onset Comparison w/ Scale against Human (\textbf{OCHuman})
    
    A one-to-one measure of similarities as in DCHuman, with an added leniency window for detecting notes.
    For each note requiring input in the human chart, if there is no corresponding note in the AI chart at that exact timestamp, we check one timestamp ahead of and behind the note.
    This metric provides a broader focus on onset detection, with the leniency window accounting for mild timing discrepancies ($\pm$23ms). 
    
    \item Overall Pattern Space (\textbf{Over. P-Space})
    
    Compares the number of unique patterns present in the generated chart versus the overall potential space of possible patterns.
    We measure patterns by using a sliding window over 8 time stamps (8 x 23ms), counting each unique ordered combination of 8 that appears.
    We use this metric to determine how much of the potential pattern space each model is covering within its output, giving a sense of novelty found within the output.
    This metric gives us insight into how much of the overall possibility space each model is covering.
    This is important because our goal is to create a model that is able to represent the large amount of variance found within Taiko charts by different authors.
    \item Human Intersection Pattern Space (\textbf{HI P-Space})
    
    Takes the intersection between the Over. P-Space of the model's chart vs the human's chart.
    This metric uses the same sliding window definition of patterning as above.
    The size of this set is compared to the size of the total set of human patterns in order to gauge what percentage of the ``human'' patterns the models are using.
\end{itemize}
With this combination of metrics, we aim to measure the performance of the models in terms of both patterning and onset detection.

\begin{table*}[]
\begin{tabular}{|c|c|c|c|}
\hline
             & \textbf{Over. P-Space} & \textbf{HI P-Space DDR} & \textbf{HI P-Space T} \\ \hline
\textbf{DDC} & 15.938\%               & 78.700\%                 & 83.160\%               \\ \hline
\textbf{TN}  & \textbf{21.328\%}      & \textbf{92.470\%}        & \textbf{94.117\%}     \\ \hline
\end{tabular}
\caption{DDC and TaikoNation compared on patterning metrics. Human Taiko dataset is abbreviated as T, and human Dance Dance Revolution dataset as DDR. Percentages are averaged from performance on all 10 charts in the dataset.}
\label{tab:pattern}
\end{table*}

\begin{table*}[]
\begin{tabular}{|c|c|c|c|c|c|}
\hline
                & \textbf{DCRand}  & \textbf{DCHuman DDR} & \textbf{DCHuman T} & \textbf{OCHuman DDR} & \textbf{OCHuman T} \\ \hline
\textbf{Random} & //                & 50.185\%            & 50.182\%          & 65.580\%            & 66.077\%          \\ \hline
\textbf{DDC}    & 49.938\%          & \textbf{76.430\%}   & \textbf{77.900\%}  & \textbf{80.900\%}   & \textbf{83.45\%}  \\ \hline
\textbf{TN}     & \textbf{50.405\%} & 74.920\%             & 74.987\%          & 79.200\%             & 79.323\%          \\ \hline
\end{tabular}
\caption{Random, DDC, and TaikoNation compared on onset detection metrics. Human Taiko dataset is abbreviated as T, and human Dance Dance Revolution dataset as DDR. Percentages are averaged from performance on all 10 charts in the dataset.}
\label{tab:onset}
\end{table*}

% the lil table for human
\begin{table*}[]
\begin{tabular}{|l|c|c|}
\hline
                     & \multicolumn{1}{l|}{\textbf{DCRand}} & \multicolumn{1}{l|}{\textbf{Over. P-Space}} \\ \hline
\textbf{Human DDR}   & \textbf{50.326\%}                       & \textbf{16.055\%}                           \\ \hline
\textbf{Human Taiko} & 50.170\%                                & 14.453\%                                    \\ \hline
\end{tabular}
\caption{Human DDR and Human Taiko datasets compared against relevant metrics. Percentages are averaged over performance on all 10 songs in the evaluation set.}
\label{tab:human}
\end{table*}

\begin{table*}[]
\begin{tabular}{|c|c|c|c|c|c|c|c|}
\hline
                     & \textbf{No Note}  & \textbf{S Don}   & \textbf{S Kat}   & \textbf{B Don}   & \textbf{B Kat}   & \textbf{Roll}    & \textbf{Denden}  \\ \hline
\textbf{Human Taiko}       & 82.180\%          & 7.394\% & 7.305\% & 0.265\%          & 0.310\%          & 0.097\%          & 2.449\% \\ \hline
\textbf{TaikoNation} & 83.270\% & 7.174\%          & 6.934\%          & 0.289\% & 0.347\% & 0.155\% & 1.830\%          \\ \hline
\end{tabular}
\caption{Distributions of notes types found in the charts for the evaluation set of human-authored and TaikoNation generated Taiko charts. The definition of each note type can be found in Section 3.2.}
\label{tab:distribution}
\end{table*}

\section{Results}
%BIG OLD TABLE (Random-> DDC -> Our approach) and columns are metrics. Bold the best value for each column. 
% the big ol table was too big and went off the page so i split it into onset detection & patterning

\begin{figure}
    \centering
    \includegraphics[width=\columnwidth]{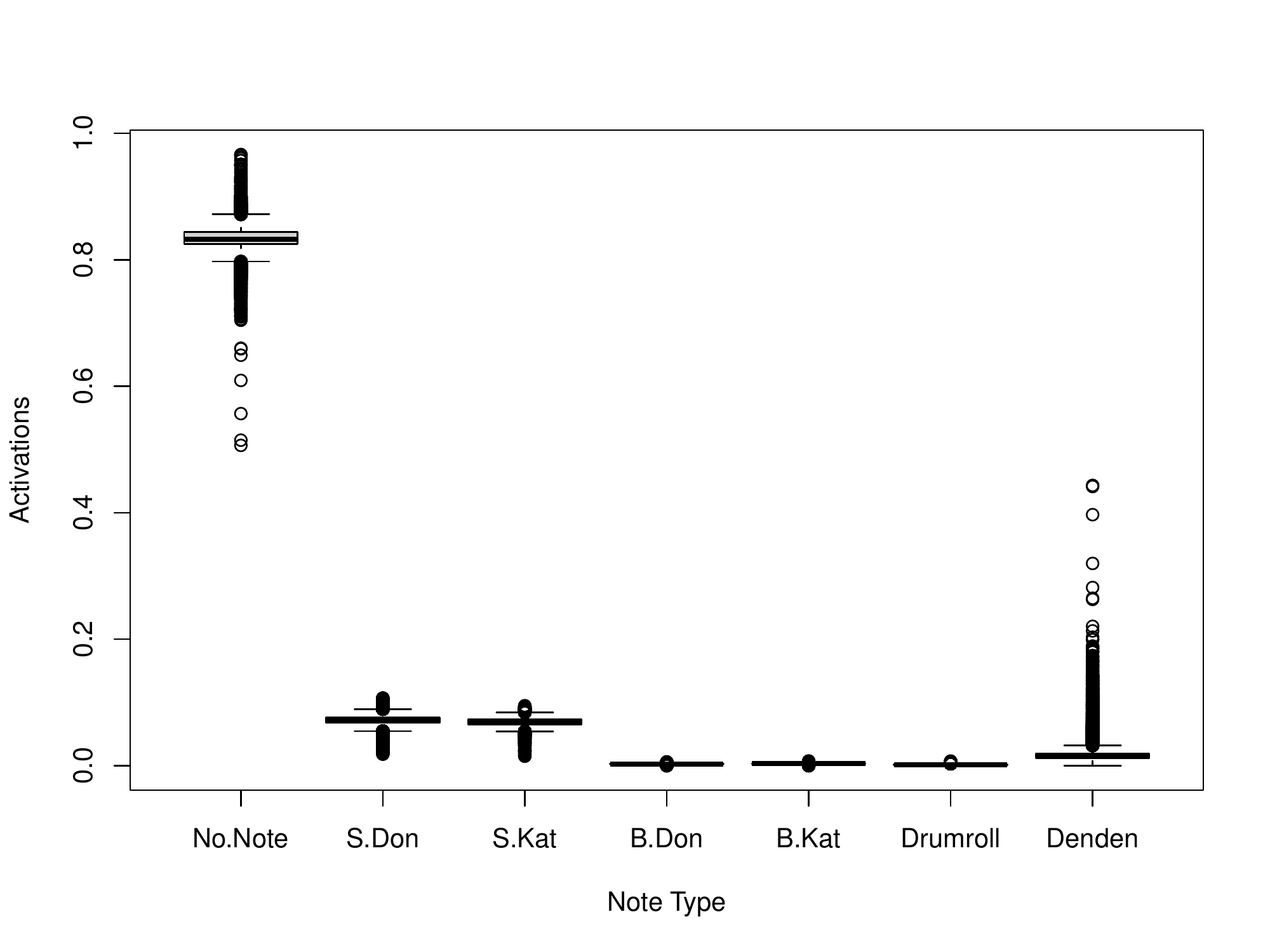}
    \caption{Boxplot depicting TaikoNation's predictions over the evaluation set. The definition of each note type can be found in Section 3.2.}
    \label{fig:pred}
\end{figure}

The results of the evaluation outlined in Section 4 can be found in Tables \ref{tab:pattern} \& \ref{tab:onset}. 
Table \ref{tab:pattern} contains the metrics measuring patterning, and Table \ref{tab:onset} contains the metrics pertaining to onset detection.
We also include Table \ref{tab:human} for additional context on how the human-authored charts performed on relevant metrics.
Table \ref{tab:distribution} and Figure \ref{fig:pred} provide information on the distribution of different note types found in the charts from the evaluation dataset.
We review these results in detail below.

%para on how we are good at patterning and what it may mean
\begin{figure*}
    \centering
    \includegraphics[width=\linewidth]{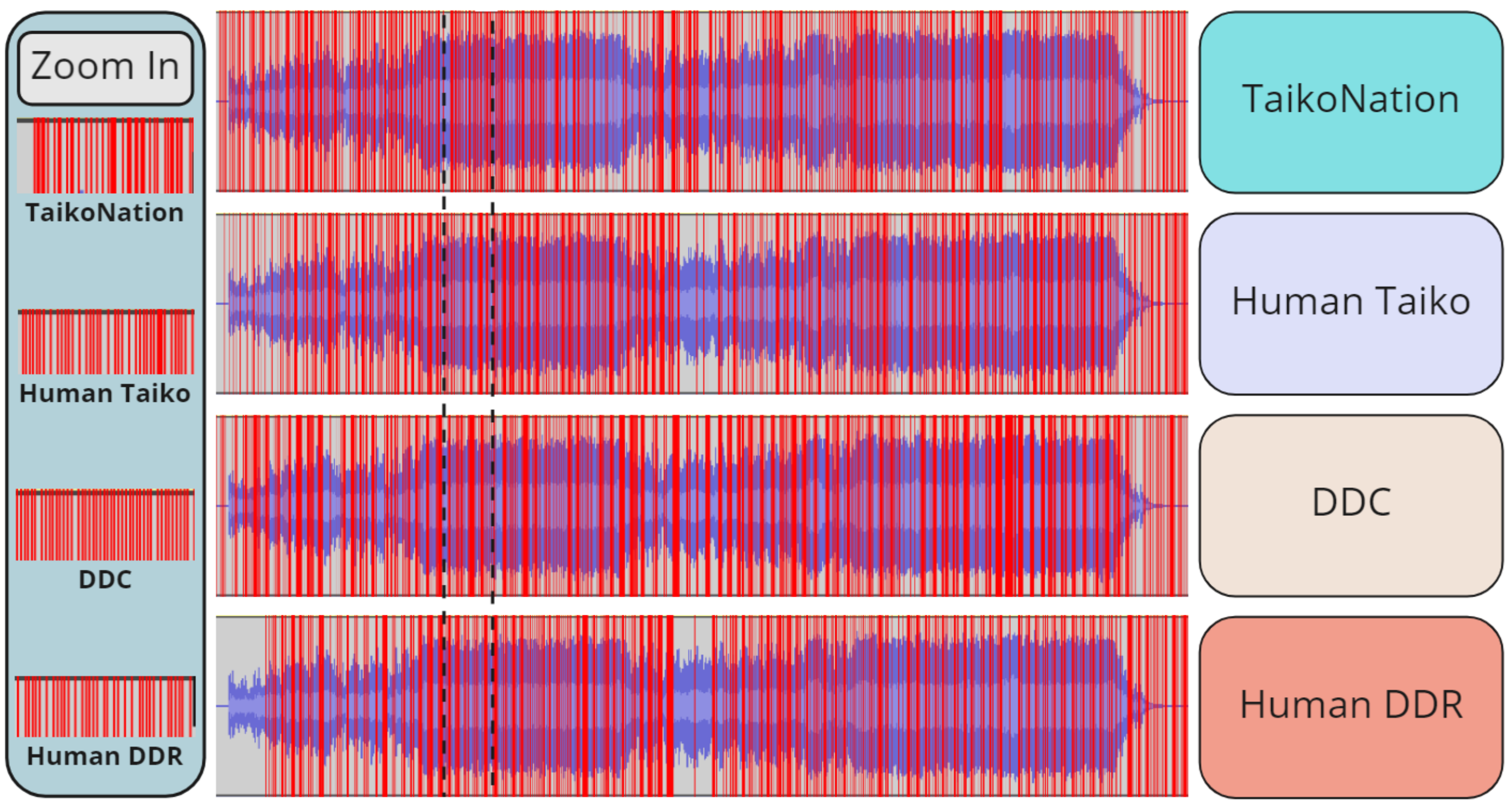}
    \caption{Model Comparison on the song \textit{Nhato - Beyond the Seven}. Left window displays a close up look at the song segment outlined in black. Each red line represents a note object.}
    \label{fig:wave}
\end{figure*}
% we use many more human patterns than DDC, DDC learning more simplistic patterns & repeating vs us using lots more human patterns, more novelty, etc (to put into words)
As shown in Table \ref{tab:pattern}, our approach (TaikoNation) uses significantly more human patterns than DDC when evaluated using both the DDR and Taiko datasets.
This means that TaikoNation is selecting many more patterns that are present in the human datasets.
In comparison, DDC is using a smaller set of patterns consistently.
A visual comparison between the two models demonstrating this difference can be seen in Figure \ref{fig:wave}, where each note is represented by a red line.
In TaikoNation's output, the density between groupings of lines is much more varied, which is closer to both the human charts for DDR and Taiko. 
Comparatively, DDC's output has a much more consistent density, which reflects our observation above.
The Overall Pattern Space (Over P-Space) for our approach is also much larger than DDC's, suggesting a larger variety of unique patterns can appear during chart generation.
We hypothesize this is partially due to the variability present in the charts in our training dataset, which were all hand-authored by separate individual authors.
% the median activation values and distributions align with the information in Table \ref{tab:distribution}.

%para on onset detection, emphasize that we are not that much behind 
As can be seen in Table \ref{tab:onset}, TaikoNation's increase of patterning did not lead to a substantial increase in similarity to random noise.
We interpret this to mean that our approach has learned more distinct human patterns without sacrificing the structure found in human charts.
Notably, both DDC and our approach performed roughly equivalently to the human datasets on the DCRand metric, as per Table \ref{tab:human}.
Likewise, our performance on the onset detection metrics was not substantially different from DDC's.
This indicates that our lack of a dedicated onset detection pipeline did not hamper our approach's onset detection ability in comparison to DDC.

%paragraph about note type selection
Notably, we are unable to compare DDC and TaikoNation across the human datasets in terms of note type selection due to the differences in game domain.
Instead we provide Figure \ref{fig:pred} and Table \ref{tab:distribution} to give insight into TaikoNation's note type selection.
Figure \ref{fig:pred} shows a boxplot of TaikoNation's activations throughout the model's predictions while generating the charts for the evaluation set.
These distributions are notably wider for more common note types such as ``no note,'' and more sparse for less common note types such as ``drumroll.''
We hypothesize this is due to the inherent class imbalance present in the dataset.
Notably, the boxplots imply that the model has picked up on the structure of one of the uncommon note types, the ``denden.''
The outliers in the denden boxplot may be occurring due to the denden's continuous structure, as it is much more likely that there is a denden object following a previous denden.
TaikoNation's ability to recognize and properly use the denden sequentially may imply the model is also capable of recreating other structural patterns of note types.
TaikoNation's ability to recognize and properly use the denden sequentially reinforces our earlier findings in Table \ref{tab:pattern} that suggest TaikoNation is using other patterns seen in human charts.
Table \ref{tab:distribution} shows that TaikoNation's generated charts have a similar distribution of each note type to the human Taiko charts in the evaluation set.
%% this would probably be a good place to add the per song distrib info if we would like to? i had a hard time wording it, let me know if you think this is sufficient
In combination with TaikoNation's performance on the patterning metrics seen in Table \ref{tab:pattern}, we note that TaikoNation is using both similar rhythmic structures as well as similar amounts of each note type when compared to human-authored charts.

%closing remark about how we did the patterns yay
Overall, our approach used more patterns found in human-authored charts than DDC without a substantial difference in onset detection metrics.
We hypothesize this is mainly due to the sliding window used for predictions, as well as the specifically curated dataset of high difficulty charts.
These results support our initial claim that our approach leads to more congruent, human-like patterning than seen in prior chart generation works.
However, we still need to confirm this with a human subject study, as our metrics are just an approximation of human judgement of patterning. 
We leave this for future work.

%Indicate all our results are in Table 1, and then walk the reader through the results and give our takes on those results.

%End with the support for our initial claim.

\section{Limitations and Future Work}

%Identify limitations of this work and what we might do to solve them
% split into tech & fundamental limitations and how to address
\subsection{Technical Limitations}
There are a few technical limitations within our approach that could be addressed to potentially improve our results.
Our initial training approach lacked robustness.
A possible solution could be training for a longer period of time with a custom stopping condition to avoid both overfitting and weight explosion.
For example, we could train until the model does not predict at least one instance of every class on a validation set to avoid under-represented classes from being ignored.
The training dataset we used was relatively small due to our focus of recreating human-like patterns found in the highest quality charts.
A possible future approach could involve training on a larger set of charts of varying qualities as a baseline, then finetuning the model with a hand-selected smaller set of higher quality charts.
Another limitation that stemmed from our focused dataset is the lack of difficulty control in our model which is present in DDC \cite{donahue2017dance}.
This was an intentional choice, as we chose to only use charts above a certain difficulty threshold in order to capture the patterning present in high difficulty charts.
It may also be worthwhile to curate a dataset created by a single author to theoretically create a more consistent model, but there are very few chart creators who are prolific enough to do so.
We did not implement more sophisticated onset detection techniques, such as the ``fuzzy labelling'' used in PCGoRG \cite{liang2019procedural}.
Implementation of these techniques within our generation pipeline could improve the onset detection of our approach.
While our division of input audio into slices of 23ms is appropriate for the vast majority of songs found in rhythm games, the system's onset detection abilities are still limited by this choice for certain edge cases.
In addition, though our pattern recognition and replication was shown to be relatively strong, they could be potentially improved through the use of architectures which have not been implemented in prior chart generation work.
For example, Transformers have been shown to perform well on sequence generation tasks \cite{dong2018speech}, and could be employed in future attempts at chart generation.

% hello: i will try to go through comments right now!
\subsection{Fundamental Limitations}
While our initial findings show that we are performing well on the task of patterning, our metrics can only provide an approximation.
In order to ascertain the value of the model, we need to conduct a human subject study to verify our findings.
There are two major types of studies that we could perform: Designer-focused and Player-focused.
Designer-focused studies would involve a group of chart creators working with our model and other baselines in a co-creative context \cite{yannakakis2014mixed}.
We imagine a co-creative study akin to that of Guzdial et al. \cite{guzdial2018co}, in which chart creators interact with our model in a mixed-initiative fashion.
%The designer could use the model how they see fit, using it to generate charts and further editing those charts by hand if they desire.
%Those designers would then give feedback about their experience using the model.
%These studies could be performed with both expert and novice creators alike, which would give us valuable insight into the potential applications and future of the approach.
Player focused studies would revolve around the comparison of human-authored charts with charts generated by our approach and other AI/ML baselines.
Players would be given both types of charts to play in a blind setting, rating specific aspects such as patterning and rhythm choices from each.
These studies would give us more insight into how well the model is learning to recreate human patterning in practice.

\subsection{Future Work}
The main hope for the future of this project is to incorporate our approach into a co-creative tool to be used by chart creators.
This tool would ideally act as a teacher to novice creators learning to chart, and a powerful tool for more experienced creators to use at their own discretion.
Building a notion of controllability into the model, meaning users could pick and choose desired aspects of the output, would be conducive to this goal.
An avenue for future work with this model could be an ablation study examining which of the new contributions were most relevant to increasing the patterning ability of TaikoNation in comparison to prior work.
This would be beneficial for future chart generation work, which could incorporate and enhance the most effective aspects of TaikoNation.
While a full ablation study was out of scope for this paper, we performed a non-exhaustive ablation study while developing the final training approach and model architecture.
The final approach used in this paper was settled on by qualitatively examining the output of the model after each change, selecting the version that produced the ``best'' patterning by our judgement.
Exploring other game domains is another possibility for future work.
There are many rhythm games with unique game objects and interesting limitations that would make the task of chart generation a difficult challenge.
For example, Sound Voltex is a rhythm game with two continuous inputs in the form of knobs which can be twisted in various ways to remix the playable songs.
This would require an approach that accounted for the keysounding and patterning of continuous inputs, which have yet to be explored.
An interesting challenge could be the development of a general architecture that could be tweaked and modified per the requirements of a particular rhythm game domain.

% moved comments down here for organization, i followed along them as i went
%possible limits: we are slightly lacking in onset detection, could we implement other techniques such as fuzzy labelling from PCGoRG or have a more sophisticated pipeline to solve this? perhaps we need a smaller time unit? not 23ms but something smaller? (10ms?) <- i realized that this would actually not help because we would have less context of the surrounding area and lose our patterning, so we'd require more data entirely for the input someone did that

%Maybe a combined pipeline is too simple and we need to use the separate onset detection & patterning pipelines, however with a more sophisticated pattern selection system? could use methods like transformers which haven't been used yet and are good for sequences

%rough approximation, but needs human testing to ascertain value 
% two major types (designer focused / player focused) in terms of how people can engage with AI gen charts! 

%additional post processing which is less simple?

%Talk at a high level about the hope for the future of this project/what we'd like to work on and why

% co-creative! other games! "controllability" improved scalability for co-creative (hey give me 4 bars of this with X rhythm)! games with more complex objects / notion of "aim" (SDVX, osu!std)
% one architecture with changes on per game basis (as baseline for future)

\section{Conclusions}

%Overview of the paper, assuming the reader has read it. Basically, if you can only have someone remember 3-4 things about this paper, what should they be?

%1. we were trying to do chart generation, specifically with focus on patterning
%2. listing the things we contributed (mirrors contributions in intro), which includes our approach, the experiments against DDC, and the dataset.
%3. broad hope for future applicability of the work 
Creating a chart generation model that replicates human-like patterning has proven to be a challenging, multifaceted task.
%Chart generation for rhythm games is a challenging, multifaceted task.
%Specifically, replicating human-like patterning within chart generation models has proven to be difficult.
We established a new approach for chart generation that produced charts with more congruent, human-like patterning than seen in prior work.
This was shown through comparisons against a leading baseline across a number of pattern-focused and onset detection metrics.
We also introduced a curated osu!Taiko dataset presented in a novel format, which could be used for a number of onset detection and chart generation tasks.
We hope to help spur future chart generation work with a stronger focus on patterning, and aid novice and expert chart creators in the chart generation process.
%This work could help spur future chart generation works, and aid creators in the chart creation process.

% the end?! unless i need to elaborate more on specifics, i did go a bit general 

\begin{acks}
%This work was funded by the Canada CIFAR AI Chairs Pro- gram. We acknowledge the support of the Natural Sciences and Engineering Research Council of Canada (NSERC).
This work was funded through a Natural Sciences and Engineering Research Council of Canada (NSERC) Undergraduate Student Research Award (USRA).
\end{acks}

\bibliographystyle{ACM-Reference-Format}
\bibliography{main}

\appendix

\end{document}